%% file: main.tex
\documentclass[runningheads]{llncs}

\usepackage{eccv}

\usepackage{eccvabbrv}

\usepackage{graphicx}
\usepackage{booktabs}

\usepackage[accsupp]{axessibility}  

\usepackage{wrapfig}
\usepackage{colortbl}
\usepackage{multirow}
\usepackage{bbding}
\usepackage{makecell}

\newcommand{\colortable}{\cellcolor[RGB]{230,230,230}}

\usepackage{hyperref}

\usepackage{orcidlink}

\begin{document}

\title{ViewFormer: Exploring Spatiotemporal Modeling \\ for Multi-View 3D Occupancy Perception \\ via View-Guided Transformers} 

\titlerunning{ViewFormer}

\author{Jinke Li\inst{1}\orcidlink{0009-0002-1958-4205} \and
Xiao He\inst{1}\orcidlink{0009-0004-6328-7552} \and
Chonghua Zhou\inst{2} \and
Xiaoqiang Cheng\inst{1}\orcidlink{0009-0002-3379-6685} \and
Yang Wen\inst{1} \and
Dan Zhang\inst{1}
}

\authorrunning{J.~Li et al.}

\institute{Uisee Foundation Research \& Development, Beijing, China \and
University of Science and Technology of China, China
}

\maketitle

\input{sec/0_abstract}
\input{sec/1_intro_v2}
\input{sec/2_relate}
\input{sec/3_method}
\input{sec/4_optimization_v3}
\input{sec/5_benchmark}
\input{sec/6_conclusion}


%
%
\bibliographystyle{splncs04}
\bibliography{main}

\input{sec/X_suppl}

\end{document}

%% file: sec/0_abstract.tex
\begin{abstract}
3D occupancy, an advanced perception technology for driving scenarios, represents the entire scene without distinguishing between foreground and background by quantifying the physical space into a grid map. The widely adopted projection-first deformable attention, efficient in transforming image features into 3D representations, encounters challenges in aggregating multi-view features due to sensor deployment constraints. To address this issue, we propose our learning-first view attention mechanism for effective multi-view feature aggregation. Moreover, we showcase the scalability of our view attention across diverse multi-view 3D tasks, including map construction and 3D object detection. Leveraging the proposed view attention as well as an additional multi-frame streaming temporal attention, we introduce ViewFormer, a vision-centric transformer-based framework for spatiotemporal feature aggregation. To further explore occupancy-level flow representation, we present FlowOcc3D, a benchmark built on top of existing high-quality datasets. Qualitative and quantitative analyses on this benchmark reveal the potential to represent fine-grained dynamic scenes. Extensive experiments show that our approach significantly outperforms prior state-of-the-art methods. The codes are available at \url{https://github.com/ViewFormerOcc/ViewFormer-Occ}.
\keywords{3D Occupancy \and Occupancy flow \and Multi-view Interaction \and Spatiotemporal Modeling \and Streaming video pipeline}

\end{abstract}

%% file: sec/1_intro_v2.tex
\section{Introduction}
\label{sec:intro}

\begin{wrapfigure}{r}{0.5\textwidth}
   \centering
    \includegraphics[width=1\linewidth]{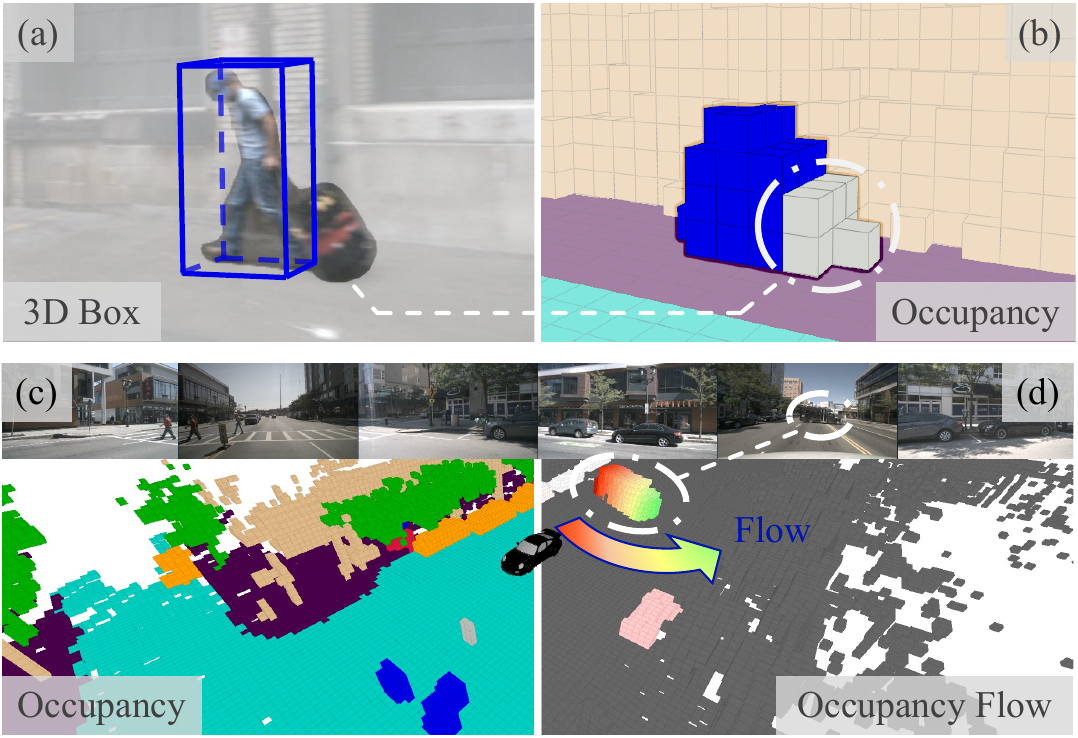}
  \caption{\small Treating objects simply as 3D boxes lacks the sense of the background, such as the suitcase in (\textbf{a}). Defining 3D space as occupancies (\textbf{b}) and (\textbf{c}) is more effective in representing objects. Beyond static occupancy, occupancy flow is crucial to perceive dynamic scenes. In the case of a turning car in (\textbf{d}), different flow directions of occupancies can be clearly observed.
  }
  \label{fig:fig1}
\end{wrapfigure}

The vision-centric autonomous driving (AD) systems are attracting extensive attention in recent years, promoting the research domain perceiving the real 3D world from 2D images. 3D object detection is a traditional task, representing foreground objects with limited categories. As illustrated in \cref{fig:fig1}(a), the pedestrian is detected as a 3D bounding box, while the uncommon suitcase is hardly defined in driving scenarios. Once such a suitcase appears on the road, an AD system relying only on object detection is unable to secure the driving safety. In contrast, the 3D occupancy representation, unifying the concept of foreground and background, and quantifying the entire 3D space into voxel-wise cells with 

\noindent semantic labels, shows superior performance, where objects like suitcases can be effectively defined as category-agnostic occupancies as shown in \cref{fig:fig1}(b). Beyond static occupancy, occupancy flow is crucial for representing dynamic scenes as shown in \cref{fig:fig1}(d). In this paper, we introduce \emph{ViewFormer}, a transformer-based framework designed to predict 3D occupancy and occupancy flow with multi-camera images as input.

Typical vision-centric 3D perception frameworks can be decomposed into spatial interaction and temporal interaction components. The former is responsible for transforming image features into 3D space, while the latter, on one hand, enhances the 3D features, and on the other hand, models dynamic scenes to reason velocity information. Regarding spatial interaction, to reduce computational costs, sparse methods have been explored to transform multi-view image features into 3D space~\cite{DBLP:conf/corl/WangGZWZ021, DBLP:conf/eccv/LiWLXSLQD22, DBLP:conf/eccv/PhilionF20}. For instance, deformable attention~\cite{DBLP:conf/iclr/ZhuSLLWD21}, originally designed for monocular images, is extended to the multi-view field by BEVFormer~\cite{DBLP:conf/eccv/LiWLXSLQD22} through projecting 3D reference points onto multiple images, which is referred to as the projection-first method in this paper. This method is widely used to aggregate multi-view features and predict 3D occupancy~\cite{sima2023_occnet, DBLP:conf/cvpr/HuangZZ0L23, Li2023FBBEV}. Nevertheless, we identify two predominant issues of this approach.

\begin{figure}[tb]
\begin{center}
\includegraphics[width=\linewidth]{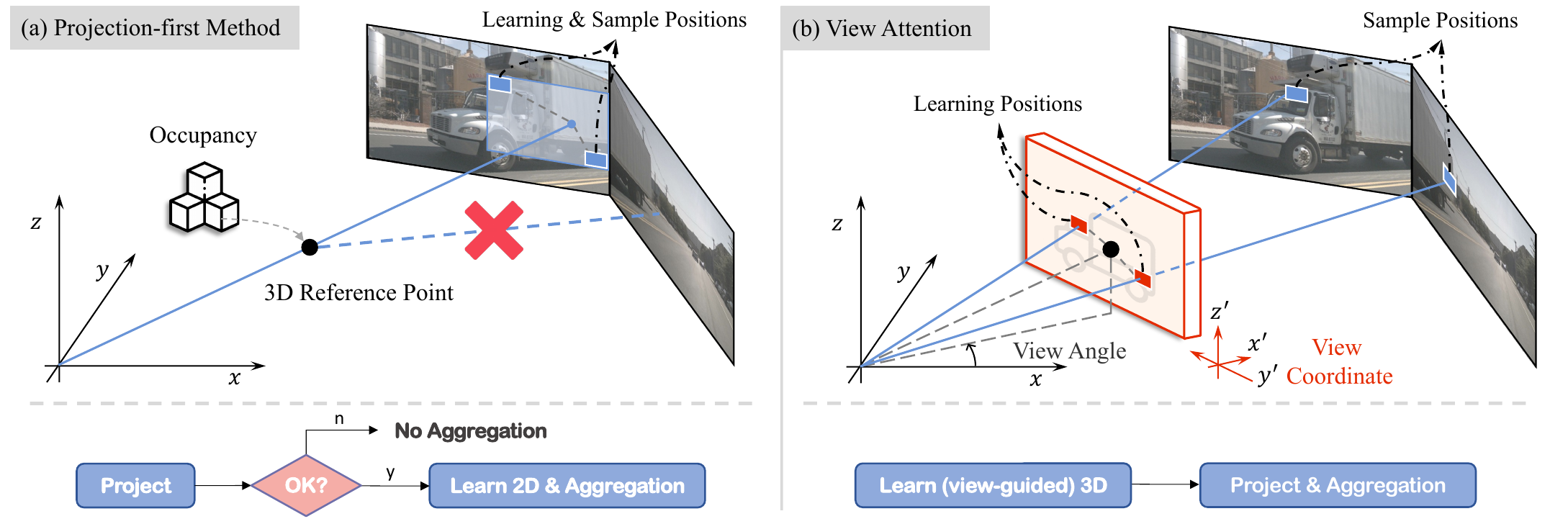}
\end{center}
   \caption{Constrained by fixed reference points, the projection-first method (\textbf{a}) introduced in \cite{DBLP:conf/eccv/LiWLXSLQD22} fails to collect multi-view features. In contrast, our learning-first view attention (\textbf{b}) gathers features from multiple cameras more adequately.}
\label{fig:fig2}
\end{figure}

To gain a better understanding, let's review the projection-first method illustrated in \cref{fig:fig2}(a), which projects the 3D reference point of a voxel query onto images first and then performs deformable attention~\cite{DBLP:conf/iclr/ZhuSLLWD21}. However, a critical limitation arises when a 3D reference point, fixed during training, is projected outside the image size for a specific camera, the projection-first method no longer applies deformable attention to extracting features for this reference point. As a consequence, features from this camera are masked out throughout the entire dataset. Notably, this issue is widespread, as seen in scenarios like nuScenes~\cite{DBLP:conf/cvpr/CaesarBLVLXKPBB20,DBLP:journals/ral/FongMHZCBV22}, where numerous 3D reference points can only be projected onto a single camera due to sensor deployment. Additionally, as the deformable attention utilized in the method learns sample points on the image plane, the 2D sample area corresponding to a 3D object undergoes rapid variations with changes in depth due to perspective transformation, such scale inconsistency poses a convergence challenge to 3D perception tasks.

To address the aforementioned issues, we introduce our learning-first {\em view attention}, facilitating multi-view feature aggregation. As shown in \cref{fig:fig2}(b), to overcome the constraint imposed by fixed reference points on feature collection, we adopt a strategy to learn local regions in 3D space for a given query. The corresponding 3D points of these regions are then projected onto multiple images for feature aggregation. Consequently, the extraction of features across cameras becomes a data-driven process. For higher efficiency and faster convergence, we define the learned 3D regions in a local view coordinate (VC) system, providing view guidance. These regions remain invariant as the query's view angle changes, introducing effective rotational invariance in perception around the vehicle. Furthermore, learning regions directly in 3D space preserves a consistent 3D spatial scale, avoiding the challenges correlated to perspective transformation.

Temporal modeling in computer vision often emphasizes the efficiency of leveraging video data~\cite{DBLP:conf/eccv/BrazilPLS20, DBLP:conf/cvpr/LuoYU18,DBLP:conf/cvpr/QiZNSVDA21}. The popular multi-camera temporal method BEVFormer~\cite{DBLP:conf/eccv/LiWLXSLQD22} achieves temporal interaction solely with a single historical frame, leading to limited performance. Besides, in BEVFormer, employing a sliding window approach during training and switching to online video during inference brings inconsistency, and the use of windowed data also increases training time due to redundant inference of many frames. Therefore, we propose our streaming temporal attention mechanism with multi-frame interaction, in which the utilization of a streaming memory mechanism~\cite{Wang_2023_ICCV, Park2022TimeWT} significantly reduces training time without additional inference latency.

In aspect of dataset, although prior work~\cite{sima2023_occnet} delves into object-level occupancy flow by assigning the center velocity of an object to all its internal occupancies, the exploration of occupancy-level flow representation remains limited. As shown in \cref{fig:fig1}(d), fine-grained occupancy flow provides more detailed information such as motion directions for different parts of a turning car. Furthermore, occupancy-level flow has the potential to represent objects whose shapes vary in motion for future research. Hence, we create our {\em FlowOcc3D} dataset, building upon nuScenes~\cite{DBLP:conf/cvpr/CaesarBLVLXKPBB20,DBLP:journals/ral/FongMHZCBV22} and Occ3D~\cite{tian2023occ3d} datasets, featuring fine-grained occupancy-level flow annotations.

In summary, our contributions are four main aspects: 1) We identify limitations of the widely used projection-first method and propose our view attention to more effectively transform multi-view features into 3D space. We demonstrate its scalability across various multi-view 3D tasks, which can be new baselines for future multi-view 3D perception research. 2) We introduce ViewFormer, a vision-centric transformer-based framework that incorporates the novel view attention and multi-frame streaming temporal attention, enchancing spatiotemporal modeling for multi-view 3D perception. 3) We create FlowOcc3D, a high-quality occupancy-level flow benchmark. Qualitative and quantitative analyses are conducted to compare models trained on occupancy-flow and object-flow, revealing the potential of fine-grained representation in dynamic scenes. 4) ViewFormer achieves state-of-the-art performance across diverse benchmarks, surpassing previous methods by substantial margins.

%% file: sec/2_relate.tex
\section{Related Work}
\label{sec:rw}

\textbf{2D Image to 3D Space Transformation.} To transform 2D image features to 3D space, bottom-up methods~\cite{DBLP:conf/eccv/PhilionF20, DBLP:conf/cvpr/ReadingHCW21, DBLP:journals/corr/abs-2206-10092} rely on pixel-wise depth prediction, where the limited reconstruction density due to corresponding image resolution poses a challenge for the demanding dense 3D occupancy representation. In contrast, top-down transformer-based methods~\cite{DBLP:conf/corl/WangGZWZ021, DBLP:conf/eccv/LiuWZS22, DBLP:conf/eccv/LiWLXSLQD22} are more flexible by allowing arbitrary predefined resolutions for 3D space and implementing feature transformation through query-to-feature interactions. For transformer-based approaches, sparse deformable attention~\cite{DBLP:conf/eccv/LiWLXSLQD22} is particularly suitable to handle a large number of occupancy instances with low computational overhead. However, we find that extending the monocular deformable attention to multi-view tasks through the projection-first method~\cite{DBLP:conf/eccv/LiWLXSLQD22}, which is also widely adopted in~\cite{sima2023_occnet, DBLP:conf/cvpr/HuangZZ0L23, Li2023FBBEV, MapTR}, presents shortcomings, motivating us to explore our view attention to adequately collect image features for multi-view AD systems.

\noindent{\textbf{3D Scene Reconstruction.}} The occupancy network, introduced by Tesla~\cite{tesla_ai_day}, brings the concept of the occupancy grid map, which has long been utilized in the domains of robotic mapping and planning~\cite{DBLP:conf/icra/SchreiberBGD21, DBLP:conf/fusion/DezertMP15}, into the AD field. Semantic scene completion~\cite{song2016ssc} is similar to the goals of occupancy tasks discussed in this paper. MonoScene~\cite{cao2022monoscene} leverages a U-Net architecture to infer dense 3D occupancy from a single image. VoxFormer~\cite{li2023voxformer} introduces depth estimation to guide voxel queries, and OccDepth~\cite{miao2023occdepth} adopts stereo depth information to improve occupancy prediction. TPVFormer~\cite{DBLP:conf/cvpr/HuangZZ0L23} proposes a tri-perspective view representation to aggregate image features. FB-OCC~\cite{Li2023FBBEV} constructs 3D features via forward-backward transformations. In addition, \cite{sima2023_occnet,wang2023openoccupancy} contribute high-quality occupancy benchmarks to facilitate the research community. Although \cite{sima2023_occnet} assigns coarse object-level velocity directly to object occupancies, fine-grained occupancy-level flow representation still remain unexplored, by which we are motivated to present our occupancy-level flow dataset FlowOcc3D, as well as a feasible framework based on our view attention and streaming temporal attention for motion 3D occupancy prediction.


%% file: sec/3_method.tex
\begin{figure}[tb]
\begin{center}
\includegraphics[width=\linewidth]{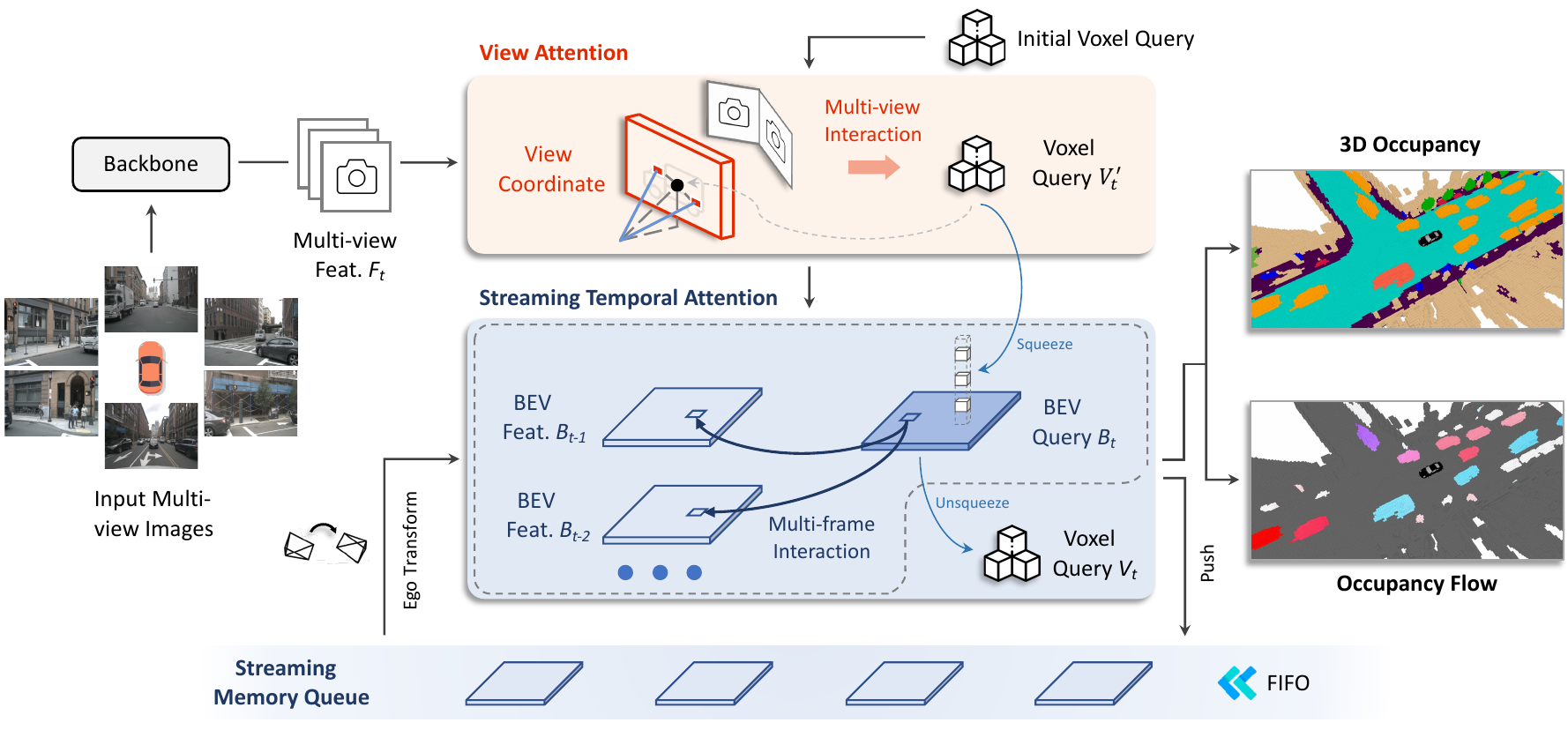}
\end{center}
   \caption{\textbf{ViewFormer pipeline.} In our ViewFormer, the multi-view features $F_t$ are first extracted from the multiple images via a backbone. Then we introduce the view attention specific for addressing the limitations of the existing projection-first method, allowing us to aggregate multi-view features for voxels $V^{\prime}_{t}$ more adequately. In our streaming temporal attention, we squeeze the voxel queries $V^{\prime}_{t}$ into the BEV queries $B_t$ with concern of the computing complexity. Each BEV cell of $B_t$ interacts with historical multi-frame BEV features stored in the streaming memory queue, where we utilize ego transformation to compensate ego motion. The voxels $V_{t}$ obtained from unsqueezing the updated BEV features are subsequently fed into 3D occupancy and occupancy flow prediction. We push the updated BEV queries into the memory queue for subsequent temporal interaction in the video stream pipeline.}
\label{fig:fig3}
\end{figure}

\section{Methodology}

In this paper, we propose a general framework, named ViewFormer, gathering spatiotemporal features from multi-view images adequately for unified prediction of 3D occupancy and occupancy flow, as shown in \cref{fig:fig3}. Our approach stands out due to two key components: the view attention and streaming temporal attention, both of which compose a transformer encoder, wherein queries circulate in the form of voxels and BEV cells, ensuring efficient computation while extracting fine-grained representative 3D features.

\noindent{\textbf{Spatial Interaction.}} Following bird’s-eye-view (BEV) perception methods~\cite{DBLP:conf/eccv/LiWLXSLQD22, DBLP:conf/corl/WangGZWZ021, DBLP:conf/eccv/LiuWZS22}, we extract multi-view image features $F_t$ via an image backbone and then update queries through spatial interaction. The difference is that we directly use voxel queries $V^{\prime}_{t}$ instead of BEV queries to collect finer-grained 3D occupancy features, where a relatively small number of query channels is set to reduce computational effort for a large number of voxel queries. Serving as the core of spatial interaction, our view attention is introduced specific for transforming multi-view image features into 3D space, exceeding a simple extension of the 2D deformable attention commonly used in~\cite{DBLP:conf/eccv/LiWLXSLQD22, sima2023_occnet, DBLP:conf/cvpr/HuangZZ0L23, Li2023FBBEV}.

\noindent{\textbf{Temporal Modeling.}} Inspired by the concept of streaming video methods~\cite{Wang_2023_ICCV, Park2022TimeWT}, we construct a streaming memory queue to dynamically store historical features spanning $N$ frames during both the training and inference phases, which follows the first-in, first-out (FIFO) principle~\cite{Wang_2023_ICCV} for entry and exit. Considering the increased burden of storage and computation, we proceed with temporal modeling at the 2D BEV level. Specifically, the voxel queries $V^{\prime}_{t}$ are squeezed into BEV-level queries $B_t$ along the z-axis, each BEV cell of $B_t$ interacts with the historical multi-frame BEV features stored in the memory queue, where we utilize ego transformation to compensate ego motion. The voxels $V_t$ obtained from unsqueezing the updated BEV queries are then fed into 3D occupancy and occupancy flow prediction. Meanwhile, we push the updated BEV queries into the memory queue for subsequent temporal interaction in the video stream pipeline.


\subsection{View Attention}
\label{sec:viewattn}

Faced with the challenge of 3D scenarios, we employ dense voxel queries ($V_t \in \mathbb{R}^{Z \times H \times W \times C_{\text{Voxel}}}$) as containers to facilitate fine-grained voxel feature extraction throughout the entire pipeline. Here $Z$, $H$, and $W$ represent the 3D dimensions of the space, and $C_{\text{Voxel}}$ denotes the feature channels. To effectively construct spatial 3D features with reasonable receptive fields in a vision-centric AD system, the keystone is to aggregate coherent 3D features from discrete multi-view images, for which we introduce our learning-first view attention mechanism to address the limitations of the existing projection-first method, as discussed \cref{sec:intro}.

Our view attention is implemented by abstracting an occupancy-related view coordinate (VC) system $\mathcal{T}$ as depicted in \cref{fig:fig2}(b). Given multi-view features $F_t$, a voxel query $\boldsymbol{q} \in \mathbb{R}^{C_{\text{Voxel}}}$ and its corresponding fixed 3D reference point $\boldsymbol{p}$, view attention can be formulated as:
\begin{equation}
    \operatorname{ViewAttn} (\boldsymbol{q}, \boldsymbol{p}, F_t) =
    \sum_{m=1}^{M} W_m \sum_{k=1}^{K} \sum_{j=1}^{J} A_{mkj} W_m^{\prime} F_{t}(\boldsymbol{p}+\Delta\boldsymbol{p}^\mathcal{T}_{mk}),
    \label{eq:eq1}
\end{equation}
where $M$, $K$ and $J$ are the number of attention heads, sample points and cameras respectively, $W_m \in \mathbb{R}^{C_{\text{Voxel}} \times C_v}$, $W_m^{\prime} \in \mathbb{R}^{C_v \times C_{\text{Voxel}}}$ are the learning weights ($C_v = C_{\text{Voxel}}/M$ by default), $A_{mkj}$ is the normalized attention weight, $\Delta\boldsymbol{p}^\mathcal{T}_{mk}$ denotes a learnable sample point associated with query $\boldsymbol{q}$ in the VC system $\mathcal{T}$, and $F_{t}(\cdot)$ represents extracting features from $F_{t}$ by projecting the learnable sample points onto multi-view images.

Now, let's delve into how we represent learnable sample points in the view-guided VC system $\mathcal{T}$. Given the coordinate $(x, y, z)$ of a reference point $\boldsymbol{p}$ for a query $\boldsymbol{q}$, we define its view angle $\theta$ as the angle between its projection line on the x-y plane and the x-axis, as illustrated in \cref{fig:fig2}(b). The VC system $\mathcal{T}$ of this query can be obtained by translating the origin of the ego-centric perception coordinate system to the reference point and rotating it around the z-axis by the view angle. The corresponding rotation matrix $\boldsymbol{R}(\theta)$ are calculated as:

\begin{equation}
\label{eq:eq2}
\begin{aligned}
\theta &=  \text{arctan2}\left(y, x \right),\\
\boldsymbol{R}(\theta) &= 
 \left[ \begin{array}{ccc}
    \cos\theta & -\sin\theta & 0\\
    \sin\theta & \cos\theta & 0\\
    0 & 0 & 1\\
  \end{array}
  \right].\\
\end{aligned}
\end{equation}

Through a transformation process, a learnable sample point $\Delta\boldsymbol{p}^\mathcal{T}$, initially defined in the local VC system $\mathcal{T}$, can be converted to be a 3D sample point $\boldsymbol{p}_{\text{s}}$ in the perception coordinate system, which is denoted as:
\begin{equation}
    \boldsymbol{p_s}
    = \boldsymbol{p}+\boldsymbol{R}(\theta)\cdot \Delta\boldsymbol{p}^\mathcal{T}.
\label{eq:eq3}
\end{equation}

Subsequently, we project all the 3D sample points associated with a query onto the multi-view images via camera projection matrices for multi-view spatial interaction, aggregating features from $F_{t}$ into voxel queries $V^{\prime}_{t}$.

\subsection{Streaming Temporal Attention}
\label{sec:streaming}

\noindent{\textbf{Streaming Memory Queue.}} Drawing inspiration from the efficient 3D point cloud perception~\cite{DBLP:conf/cvpr/LiHWGCZ22, DBLP:conf/cvpr/LangVCZYB19}, where it is observed that fine-grained 3D feature representation is not necessarily required for 3D space, we recognize that combining the voxel level and the BEV level can significantly enhance computation efficiency and alleviate storage consumption. Therefore, we establish our BEV-level streaming memory queue $\boldsymbol{B} = \{B_i \in \mathbb{R}^{H \times W \times C_{\text{BEV}}}, i = t-1, ..., t-N\}$ with $C_{\text{BEV}}$ $\textgreater$ $C_{\text{Voxel}}$, to dynamically store historical features spanning $N$ frames during both the training and inference phases. The latest BEV features are pushed into the memory queue per frame for later BEV-level temporal interaction.

\noindent{\textbf{Multi-frame Temporal Interaction.}} We perform a multi-frame streaming temporal interaction. We observe that, compared to voxel queries $V^{\prime}_{t}$ directly interacting with all historical BEV features in the memory queue, implementing BEV-level temporal interaction not only reduces computational overhead but also improves accuracy. Specifically, we first compress the voxel queries $V^{\prime}_{t}$ into BEV queries $B_t \in \mathbb{R}^{H \times W \times C_{\text{BEV}}}$ along the z-axis, of which each BEV cell interacts with all the memory BEV features by leveraging the deformable attention~\cite{DBLP:conf/iclr/ZhuSLLWD21}, and the updated BEV queries are then recovered to be the voxel queries $V_t$ through a feed-forward function for final predictions. We attribute this improvement to the presence of a substantial number of empty voxels, the compressed BEV queries $B_t$ lead to more pure information and thus enhance the temporal interaction.

To compensate ego motion, we apply feature warping by transforming all memory BEV features to the current frame as in~\cite{huang2022bevdet4d, Park2022TimeWT, DBLP:journals/corr/abs-2206-10092}. Taking the last frame $t-1$ as an example, given the ego pose matrices of the last frame $T_{t-1}$ and the current frame $T_t$, the transformation matrix $T_{t-1}^t$ between two frames is calculated as follow:

\begin{equation}
    T_{t-1}^t = T_t^{inv}\cdot\ T_{t-1}.
\label{eq:eq4}
\end{equation}

We align the last BEV features $B_{t-1}$ to the current frame:
\begin{equation}
    \tilde{B}_{t} = T_{t-1}^t\cdot\ B_{t-1},
\label{eq:eq5}
\end{equation}
where $\tilde{B}_{t}$ is the aligned BEV features in the local coordinate system of the current ego pose. Then, we adopt deformable attention~\cite{DBLP:conf/iclr/ZhuSLLWD21} to achieve interaction between the current BEV queries $B_t$ and the aligned multi-frame BEV features $\boldsymbol{\tilde{B}_{t}}$ for the final updated BEV queries. The multi-frame aggregation mechanism follows a multi-scale aggregation~\cite{DBLP:conf/iclr/ZhuSLLWD21}.

%% file: sec/4_optimization_v3.tex
\section{Optimization}
\subsection{Occupancy Flow Generation}

\begin{wrapfigure}{r}{0.5\textwidth}
   \centering
    \includegraphics[width=0.5\textwidth]{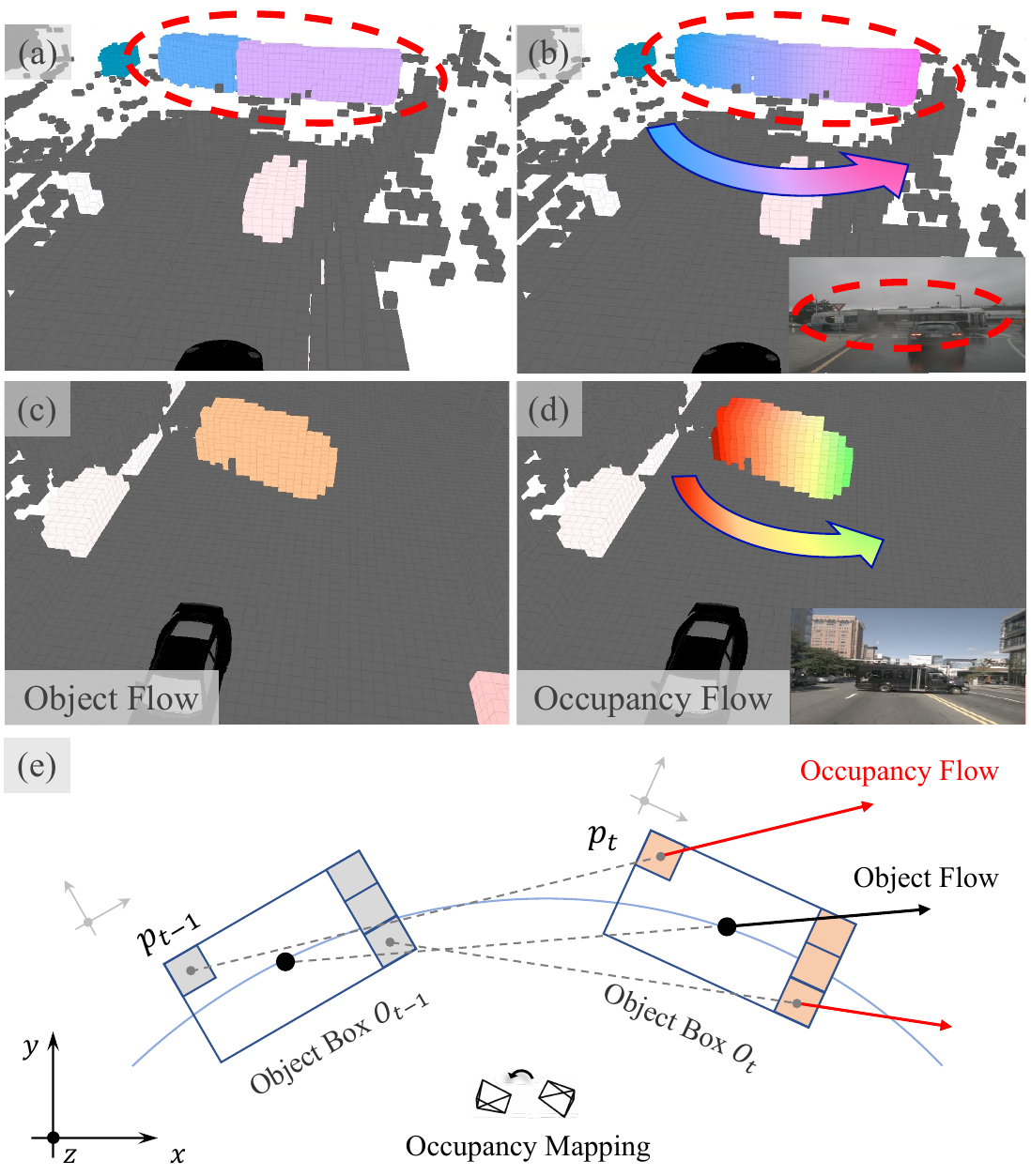}
  \caption{\small \textbf{Occupancy flow vs. object flow.} Object flow assigns only a single flow vector to the entire object as in (\textbf{a}) and (\textbf{c}), while occupancy flow provides finer-grained flow vectors for all occupancy grids as in (\textbf{b}) and (\textbf{d}), where the color and brightness represent the flow direction and magnitude respectively.
  }
  \label{fig:flowanno}
\end{wrapfigure}

Although OpenOcc~\cite{sima2023_occnet} has already generated object-level flow annotations by assigning the center velocity of an object to all its internal occupancies, the occupancy-level flow representation still remains unexplored. As illustrated in \cref{fig:flowanno}, fine-grained occupancy flow is able to capture more accurate flow vectors for different parts of a turning vehicle, offering the potential for a more precise representation of dynamic scenes. This capability is beneficial for decision-making in an AD system. Hence, we build our FlowOcc3D dataset with occupancy-level flow annotations. \cref{fig:flowanno}(e) illustrates the generation process of flow annotations. As nuScenes~\cite{DBLP:conf/cvpr/CaesarBLVLXKPBB20,DBLP:journals/ral/FongMHZCBV22} does not provide explicit temporal association for occupancies themselves, we leverage the temporal association of objects to track their internal occupancies. Given the pose matrices $O_t$ and $O_{t-1}$ of an object in two frames, along with an occupancy center $p_t$ within the object box at $O_t$, where $O_t$, $O_{t-1}$ and $p_t$ are all defined in the global coordinate system, by taking the box as a rigid body, we map $p_t$ to obtain its historical position $p_{t-1}$ in the temporal frame ${t-1}$, via ${p}_{t-1} = O_{t-1} \cdot O^{inv}_t \cdot p_t$. Then, the corresponding flow vector can be computed by $f = (p_t - {p}_{t-1}) / \Delta t$, where $\Delta t$ denotes the time interval.

\subsection{Loss Function}
Our final loss function is comprised of four parts, the focal loss~\cite{DBLP:conf/iccv/LinGGHD17} $\mathcal{L}_{focal}$ for supervising occupancy state, the cross-entropy loss $\mathcal{L}_{ce}$ as well as the Lovasz softmax loss \cite{DBLP:conf/cvpr/BermanTB18} $\mathcal{L}_{ls}$ for semantic classification, and the L1 loss $\mathcal{L}_{l1}$ with $\lambda$ to adjust loss weight for flow regression, which can be formulated as:

\begin{equation}
\mathcal{L}=\mathcal{L}_{focal}+\mathcal{L}_{ce}+\mathcal{L}_{ls}+\lambda\mathcal{L}_{l1}.
\end{equation}

\subsection{Implementation Details}

\noindent\textbf{Network.} Following the experimental setup \cite{DBLP:conf/eccv/LiWLXSLQD22, sima2023_occnet}, we employ two backbones: ResNet50 (Res50)~\cite{he2016resnet} initialized from ImageNet~\cite{deng2009imagenet}, and ResNet101 (Res101)~\cite{he2016resnet} initialized from FCOS3D \cite{wang2021fcos3d} for experiments in \cref{tab:occ_method_compe} and \cref{tab:occ_method}. In respect of experiments in \cref{tab:occ_vel_compe} and the ablation section, we utilize InternImage-Tiny (InterT)~\cite{Wang_2023_CVPR} initialized from COCO~\cite{DBLP:conf/eccv/LinMBHPRDZ14} as image backbone. Our transformer encoder has 4 layers. For the detection space of the voxel queries $V_t$, we define the dimensions as: $H=100$, $W=100$, $Z=8$, and $C_{\text{Voxel}}=72$, while the channels of the BEV queries $B_t$ are set to $C_{\text{BEV}}=126$. We use $N=4$ temporal frames for the streaming memory queue.

\noindent\textbf{Training and Inference.} By default, we train our ViewFormer with a streaming video approach \cite{Wang_2023_ICCV, Park2022TimeWT} for 24 epochs. The learning rate is set to $2\times 10^{-4}$, and the image size is $256 \times 704$. We also apply interpolation method on predictions to align with ground truth whose resolutions are inconsistent with our detection space. For the ease of comparison among different methods, we predict and evaluate BEV-level flow, and for visualization, we map the BEV-level flow to each voxel cell to recover occupancy flow. Inference FPS is measured with a single RTX 3090 GPU, while the training time is recorded using 8 A100 GPUs.

%% file: sec/5_benchmark.tex
\section{Evaluation}

To evaluate the performance of our method comprehensively, we utilize the high-quality 3D occupancy benchmarks Occ3D~\cite{tian2023occ3d} and OpenOcc~\cite{sima2023_occnet}. Moreover, based on the nuScenes~\cite{DBLP:conf/cvpr/CaesarBLVLXKPBB20,DBLP:journals/ral/FongMHZCBV22} and Occ3D~\cite{tian2023occ3d} datasets, we build our FlowOcc3D with occupancy-level flow annotations to study the potential of fine-grained occupancy flow representation, on which we further conduct extensive experiments.

\subsection{Datasets}

Both of the Occ3D~\cite{tian2023occ3d} and OpenOcc~\cite{sima2023_occnet} occupancy datasets are derived from nuScenes~\cite{DBLP:conf/cvpr/CaesarBLVLXKPBB20,DBLP:journals/ral/FongMHZCBV22}, a substantial driving dataset that comprises videos of 1000 scenes equipped with multi-view cameras. It includes 34,149 annotated frames for all 700 training scenes and 150 validation scenes.

\noindent\textbf{Benchmark Details.} \textbf{Occ3D}~\cite{tian2023occ3d} defines the detection space $\mathcal{V}=[-40\text{m}, 40\text{m}] \times [-40\text{m}, 40\text{m}] \times [-1\text{m}, 5.4\text{m}]$, in the ego vehicle coordinate system to generate occupancy data. The space $\mathcal{V}$ is voxelized with a resolution of $\Delta s=0.4\text{m}$, producing $200 \times 200 \times 16$ voxel grids to represent the 3D environment. Furthermore, the dataset includes occupancy visibility masks for various sensors, facilitating performance improvement and evaluation in diverse tasks. To create our \textbf{FlowOcc3D} dataset, we attach the occupancy-level flow information to the occupancy cells of Occ3D~\cite{tian2023occ3d} for each annotated frame. \textbf{OpenOcc}~\cite{sima2023_occnet} defines the detection space $\mathcal{V}=[-50\text{m}, 50\text{m}] \times [-50\text{m}, 50\text{m}] \times [-5\text{m}, 3\text{m}]$ in LiDAR coordinate system, and 
produces $200 \times 200 \times 16$ voxel grids by voxelizing $\mathcal{V}$ with a resolution of $\Delta s=0.5\text{m}$.


\noindent\textbf{Evaluation Metrics.} Occ3D utilizes the mean Intersection-over-Union (mIoU) across categories. OpenOcc uses both the mIoU and the single-class $\text{IoU}_{geo}$ as metrics to evaluate the occupancy state. Besides these two metrics, we additionally compute the mean absolute velocity error (mAVE) across categories to evaluate occupancy flow on our FlowOcc3D benchmark.


\begin{table*}[t]
\caption{\textbf{3D Occupancy Prediction on Occ3D benchmark.} Our ViewFormer outperforms previous SOTAs by significant margins.
}
\small
\setlength{\tabcolsep}{3.5pt}
\centering
\resizebox{\textwidth}{!}{

\begin{tabular}{@{}l|c|>{\colortable}c|ccccccccccccccccc@{}}
\toprule
Method                  & Backbone  & mIoU  &\rotatebox{90}{others} &\rotatebox{90}{barrier} & \rotatebox{90}{bicycle} &  \rotatebox{90}{bus}   & \rotatebox{90}{car}   & \rotatebox{90}{const. veh.} & \rotatebox{90}{motorcycle} & \rotatebox{90}{pedestrian} & \rotatebox{90}{traffic cone} & \rotatebox{90}{trailer} & \rotatebox{90}{truck} & \rotatebox{90}{driv. surf.} & \rotatebox{90}{other flat} & \rotatebox{90}{sidewalk} & \rotatebox{90}{terrain} & \rotatebox{90}{manmade} & \rotatebox{90}{vegetation} \\
\midrule
MonoScene~\cite{cao2022monoscene}              & Res101 & 6.06 & 1.75 & 7.23 & 4.26 & 4.93 & 9.38 & 5.67 & 3.98 & 3.01 & 5.90 & 4.45 & 7.17 & 14.91 & 6.32 & 7.92 & 7.43 & 1.01 & 7.65      \\
TPVFormer~\cite{DBLP:conf/cvpr/HuangZZ0L23}  & Res101 & 27.83 & 7.22 & 38.90 & 13.67 & 40.78 & 45.90 & 17.23 & 19.99 & 18.85 & 14.30 & 26.69 & 34.17 & 55.65 & 35.47 & 37.55 & 30.70 & 19.40 & 16.78 \\
OccFormer~\cite{zhang2023occformer}  & Res101 & 21.93 & 5.94 & 30.29 & 12.32 & 34.40 & 39.17 & 14.44 & 16.45 & 17.22 & 9.27 & 13.90 & 26.36 & 50.99 & 30.96 & 34.66 & 22.73 & 6.76 & 6.97 \\
BEVFormer~\cite{DBLP:conf/eccv/LiWLXSLQD22}  & Res101 & 26.88 & 5.85 & 37.83 & 17.87 & 40.44 & 42.43 & 7.36 & 23.88 & 21.81 & 20.98 & 22.38 & 30.70 & 55.35 & 28.36 & 36.00 & 28.06 & 20.04 & 17.69 \\
CTF-Occ~\cite{tian2023occ3d}  & Res101 & 28.53 & 8.09 & 39.33 & 20.56 & 38.29 & 42.24 & 16.93 & 24.52 & 22.72 & 21.05 & 22.98 & 31.11 & 53.33 & 33.84 & 37.98 & 33.23 & 20.79 & 18.00 \\
\midrule
FB-OCC~\cite{Li2023FBBEV}  & Res50 & 39.11 & \textbf{13.57} & 44.74 & 27.01 & \textbf{45.41} & 49.10 & \textbf{25.15} & 26.33 & 27.86 & 27.79 & 32.28 & 36.75 & 80.07 & 42.76 & 51.18 & 55.13 & 42.19 & 37.53 \\
\rowcolor[RGB]{230,230,230}ViewFormer (ours) & Res50 & \textbf{41.85} & 12.94 & \textbf{50.11} & \textbf{27.97} & 44.61 & \textbf{52.85} & 22.38 & \textbf{29.62} & \textbf{28.01} & \textbf{29.28} & \textbf{35.18} & \textbf{39.40} & \textbf{84.71} & \textbf{49.39} & \textbf{57.44} & \textbf{59.69} & \textbf{47.37} & \textbf{40.56} \\
\bottomrule
\end{tabular}
}
\label{tab:occ_method_compe}
\end{table*}

\begin{table*}[t]
\caption{\textbf{3D Occupancy Prediction on OpenOcc benchmark.} Our ViewFormer demonstrates significant performance improvements over previous SOTAs in terms of both the mIoU and IoU$_{geo}$.
}
\small
\setlength{\tabcolsep}{3.5pt}
\centering
\resizebox{\textwidth}{!}{

\begin{tabular}{@{}l|c|>{\colortable}c|c|cccccccccccccccc@{}}
\toprule
Method                  & Backbone  & mIoU & \text{IoU}$_{geo}$  &\rotatebox{90}{barrier} & \rotatebox{90}{bicycle} &  \rotatebox{90}{bus}   & \rotatebox{90}{car}   & \rotatebox{90}{const. veh.} & \rotatebox{90}{motorcycle} & \rotatebox{90}{pedestrian} & \rotatebox{90}{traffic cone} & \rotatebox{90}{trailer} & \rotatebox{90}{truck} & \rotatebox{90}{driv. surf.} & \rotatebox{90}{other flat} & \rotatebox{90}{sidewalk} & \rotatebox{90}{terrain} & \rotatebox{90}{manmade} & \rotatebox{90}{vegetation} \\
\midrule
TPVFormer~\cite{DBLP:conf/cvpr/HuangZZ0L23}              & Res101 & 23.67 & 37.47 & 27.95   & 12.75   & 33.24 & \textbf{38.70}  & 12.41                 & 17.84      & 11.65      & 8.49          & \textbf{16.42}   & 26.47 & 47.88              & 25.43       & 30.62    & 30.18   & 15.51   & 23.12      \\
OccNet~\cite{sima2023_occnet} & Res101 & 26.98 & 41.08 & \textbf{29.77}   & 16.89   & 34.16 & 37.35 & 15.58                 & \textbf{21.92}      & \textbf{21.29}      & \textbf{16.75}         & 16.37   & 26.23 & 50.74              & 27.93       & 31.98    & 33.24   & 20.80    & 30.68 \\
\rowcolor[RGB]{230,230,230}ViewFormer (ours)  & Res101 & \textbf{27.37} & \textbf{41.86} & 28.18   & \textbf{17.96}   & \textbf{34.49} & 37.03 & \textbf{16.00}                 & 21.53      & 19.50      & 13.56         & 15.59   & \textbf{26.52} & \textbf{51.48}              & \textbf{31.81}       & \textbf{34.73}    & \textbf{34.77}   & \textbf{22.20}    & \textbf{32.62}     \\
\midrule

BEVDet4D \cite{huang2022bevdet4d}                 & Res50  & 9.85 & 18.27  & 13.56   & 0.00       & 13.04 & 26.98 & 0.61                  & 1.20        & 6.76       & 0.93          & 1.93    & 12.63 & 27.23              & 11.09       & 13.64    & 12.04   & 6.42    & 9.56       \\
BEVDepth \cite{DBLP:journals/corr/abs-2206-10092}               & Res50  & 11.88 & 23.45  & 15.15   & 0.02    & 20.75 & 27.05 & 1.10                   & 2.01       & 9.69       & 1.45          & 1.91    & 14.31 & 31.92              & 7.88        & 17.08    & 16.27   & 8.76    & 14.75      \\
BEVDet \cite{huang2021bevdet}                & Res50  & 12.49 & 27.46 & 16.06   & 0.11    & 18.27 & 21.09 & 2.62                  & 1.42       & 7.78       & 1.08          & 3.4     & 13.76 & 33.89              & 10.84       & 17.55    & 22.03   & 11.72   & 18.15      \\
OccNet~\cite{sima2023_occnet}  & Res50 & 19.48 & 37.69  & 20.63   & 5.52   & 24.16 & 27.72 & 9.79                 & 7.73      & 13.38      & 7.18         & 10.68   & 18.00 &  46.13              & 20.6        & 26.75    & 29.37   & 16.90    & 27.21 \\

\rowcolor[RGB]{230,230,230}ViewFormer (ours)  & Res50 & \textbf{23.84} & \textbf{40.40} & \textbf{24.35}   & \textbf{12.55}   & \textbf{28.87} & \textbf{31.87} & \textbf{15.35}                 & \textbf{17.89}      & \textbf{14.76}      & \textbf{8.53}         & \textbf{14.18}   & \textbf{23.54} &  \textbf{49.02}              & \textbf{29.05}        & \textbf{32.35}    & \textbf{32.16}   & \textbf{17.93}    & \textbf{29.00}\\
\bottomrule
\end{tabular}
}
\label{tab:occ_method}
\end{table*}

\begin{table*}[t]
\caption{\textbf{3D Occupancy and Occupancy Flow Prediction on FlowOcc3D.} Our ViewFormer achieves the best result among previous SOTAs. *: we add a flow head for flow prediction to BEVFormer~\cite{DBLP:conf/eccv/LiWLXSLQD22} and FB-OCC~\cite{Li2023FBBEV} and retrain them on FlowOcc3D.
}
\small
\setlength{\tabcolsep}{2.0pt}
\centering
\resizebox{\textwidth}{!}{

\begin{tabular}{@{}l|c|c|>{\colortable}c|c|c|ccccccccccccccccc@{}}
\toprule
Method                  & Backbone & \makecell{Flow \\ Head} & mIoU & \text{IoU}$_{geo}$   & mAVE$\downarrow$  &\rotatebox{90}{others} &\rotatebox{90}{barrier} & \rotatebox{90}{bicycle} &  \rotatebox{90}{bus}   & \rotatebox{90}{car}   & \rotatebox{90}{const. veh.} & \rotatebox{90}{motorcycle} & \rotatebox{90}{pedestrian} & \rotatebox{90}{traffic cone} & \rotatebox{90}{trailer} & \rotatebox{90}{truck} & \rotatebox{90}{driv. surf.} & \rotatebox{90}{other flat} & \rotatebox{90}{sidewalk} & \rotatebox{90}{terrain} & \rotatebox{90}{manmade} & \rotatebox{90}{vegetation} \\
\midrule
BEVFormer*~\cite{DBLP:conf/eccv/LiWLXSLQD22}  & InternT & \Checkmark & 33.61 & 67.41 & 0.695 & 8.44 & 40.80 & 12.57 & 37.70 & 45.76 & 18.44 & 12.14 & 22.52 & 21.25 & 25.97 & 29.40 & 80.92 & 38.19 & 48.56 & 52.58 & 40.98 & 35.07 \\
FB-OCC*~\cite{Li2023FBBEV}  & InternT & \Checkmark & 37.36 & 69.73 & 0.433 & 10.95 & 40.08 & 23.14 & 42.87 & 47.17 & \textbf{21.43} & 23.84 & 27.24 & 24.50 & 31.54 & 37.36 & 80.31 & 42.26 & 50.14 & 55.44 & 39.98 & 36.84 \\
\rowcolor[RGB]{230,230,230}ViewFormer (ours) & InternT & \Checkmark & \textbf{42.54} & \textbf{72.36} & \textbf{0.412} & \textbf{13.63} & \textbf{49.35} & \textbf{28.74} & \textbf{46.63} & \textbf{52.71} & 21.04 & \textbf{29.63} & \textbf{30.34} & \textbf{30.53} & \textbf{34.01} & \textbf{41.04} & \textbf{85.23} & \textbf{50.63} & \textbf{58.68} & \textbf{61.63} & \textbf{47.72} & \textbf{41.56} \\
\midrule
BEVFormer~\cite{DBLP:conf/eccv/LiWLXSLQD22} & InternT & \XSolidBrush & 36.93 & 68.49 & - & 8.56 & 42.94 & 19.34 & 47.02 & 49.59 & 20.36 & 22.62 & 24.69 & 19.77 & 30.11 & 35.24 & 82.11 & 40.86 & 50.62 & 54.81 & 42.56 & 36.55 \\
FB-OCC~\cite{Li2023FBBEV} & InternT & \XSolidBrush & 38.69 & 69.95 & - & 11.4 & 41.42 & 24.27 & 46.01 & 49.38 & \textbf{24.56} & 27.06 & 28.09 & 25.61 & 32.23 & 38.46 & 80.97 & 42.99 & 50.95 & 56.15 & 40.55 & 37.61 \\
\rowcolor[RGB]{230,230,230}ViewFormer (ours) & InternT & \XSolidBrush & \textbf{43.61} & \textbf{72.46} & - & \textbf{13.82} & \textbf{50.32} & \textbf{29.49} & \textbf{49.24} & \textbf{54.52} & 24.34 & \textbf{32.72} & \textbf{31.09} & \textbf{31.49} & \textbf{34.44} & \textbf{41.62} & \textbf{85.47} & \textbf{51.27} & \textbf{59.03} & \textbf{62.15} & \textbf{48.33} & \textbf{42.06} \\
\bottomrule
\end{tabular}
}
\label{tab:occ_vel_compe}
\end{table*}

\subsection{Main Results}
\label{sec:mainRS}

\textbf{3D Occupancy on Occ3D.} We compare our ViewFormer with previous state-of-the-art methods on the 3D occupancy task in \cref{tab:occ_method_compe}, where the baselines BEVFormer~\cite{DBLP:conf/eccv/LiWLXSLQD22}, TPVFormer~\cite{DBLP:conf/cvpr/HuangZZ0L23}, CTF-Occ~\cite{zhang2023occformer} and FB-OCC~\cite{Li2023FBBEV} adopt the projection-first spatial interaction method as analyzed in \cref{sec:intro}. In terms of training setup, we follow FB-OCC~\cite{Li2023FBBEV}, the winner of 3D occupancy challenge of CVPR 2023~\cite{tian2023occ3d, sima2023_occnet}, which includes 90-epoch training, the same pretrained weights, and depth supervision for image backbone. As shown in \cref{tab:occ_method_compe}, in contrast to the strong baseline FB-OCC~\cite{Li2023FBBEV}, which utilizes longer video sequences, our model achieves 2.74 mIoU improvement, proving its superiority.

\noindent{\textbf{3D Occupancy on OpenOcc.}} We conduct experiments on OpenOcc in \cref{tab:occ_method}, where we adopt the same configuration as OccNet~\cite{sima2023_occnet}, with an image size of 450 $\times$ 800. Our approach demonstrates significant performance improvements over the baselines. Especially, we surpass the state-of-the-art OccNet~\cite{sima2023_occnet} by 4.36 on mIoU and 2.71 on $\text{IoU}_{geo}$ when utilizing Res50 as the backbone.

\noindent{\textbf{Occupancy and Flow on FlowOcc3D.}} We validate our ViewFormer against previous state-of-the-art methods on the 3D occupancy and flow tasks with our FlowOcc3D in \cref{tab:occ_vel_compe}, training both the occupancy and flow heads simultaneously. The two selected baselines, BEVFormer~\cite{DBLP:conf/eccv/LiWLXSLQD22} and FB-OCC~\cite{Li2023FBBEV}, employ two typical temporal modeling approaches that can directly affect the flow prediction. The former utilizes a transformer-based single-frame interaction trained in a sliding-window fashion, while the latter adopts a CNN-based streaming video interaction from~\cite{Park2022TimeWT}. We add a flow head to the two methods and retrain them on our FlowOcc3D for 24 epochs with an image size of 256 $\times$ 704, matching our ViewFormer setup. The results in \cref{tab:occ_vel_compe} show that our method surpasses the two baselines in both the occupancy and flow prediction. It can be observed that, under the 24-epoch setup, our model outperforms FB-OCC by 5.2 mIoU, 2.63 $\text{IoU}_{geo}$ and 2.1\% mAVE, highlighting the rapid convergence of our method. We also report the results without the flow head under the same settings.

\subsection{Ablations and Analysis}

\noindent{\textbf{View Attention.}} To validate the effectiveness of our view attention, we conduct experiments as presented in \cref{tab:TPattn}. Since this module serves as a spatial interaction and is not involved in temporal modeling, we temporarily disable the flow head. The findings are as below. 1) We replace our view attention module with the projection-first deformable attention as BEVformer~\cite{DBLP:conf/eccv/LiWLXSLQD22}, the results show that our learning-first view attention leads to an improvement of 1.22 mIoU and 0.43 IoU$_{geo}$ under approximate computational complexity, revealing the limitation of the commonly used projection-first method for multi-view feature aggregation. 2) Learning sample points directly in the ego-centric perception coordinate system instead of the query-specific view coordinate (VC) system results in performance degradation of -0.75 mIoU, which validates our motivation mentioned in \cref{sec:intro} that the rotation invariance introduced by our view attention effectively reduces learning complexity, accelerates convergence, and enhances accuracy. Detailed convergence experiments can be found in supplementary materials.

\begin{table}[t]
\caption{\textbf{Ablation study for the view attention.} ``Projection-first'' denotes the extension of deformable attention used in the multi-view field as ~\cite{DBLP:conf/eccv/LiWLXSLQD22}. ``w/o VC trans.'' denotes learning sample points directly in the ego-centric perception coordinate system without view angle rotation.}
\setlength{\tabcolsep}{3.5pt}
\centering
\begin{tabular}
{l|>{\colortable}c|c|ccc}
\toprule
Method & mIoU & \text{IoU}$_{geo}$ & car & truck & vege. \\
\midrule
View Attention & \textbf{43.61} & \textbf{72.46}  & \textbf{54.52} & \textbf{41.62} & \textbf{42.06} \\
Projection-first & 42.39 & 72.03 & 53.16 & 40.58 & 40.99 \\
\midrule
w/o VC trans. & 42.86 & 72.26  & 53.30 & 40.70 & 41.93\\
\bottomrule
\end{tabular}
\label{tab:TPattn}
\end{table}

\begin{wrapfigure}{r}{0.5\textwidth}
   \centering
    \includegraphics[width=1.0\linewidth]{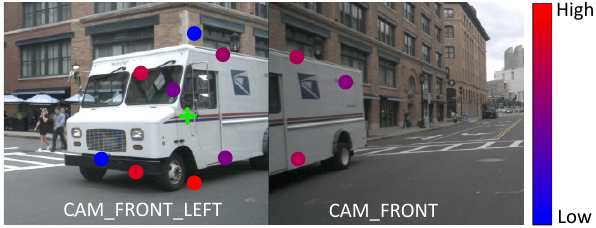}
  \caption{\small \textbf{Visualization on ViewAttn..}
  }
  \label{fig:vva}
\end{wrapfigure}

\noindent{\textbf{Visualization Analysis.}} We visualize our view attention in \cref{fig:vva}. The green cross mark denotes a query's reference point, which can only be projected onto the left image. The learned 3D points are projected onto multi-view images represented by filled circles, with colors to indicate attention weights. In such scenarios, the projection-first method, like BEVFormer~\cite{DBLP:conf/eccv/LiWLXSLQD22} that only gathers features from the image where the reference point can be projected, fails to collect features from the right image, as discussed in \cref{sec:intro}. In contrast, our method is competent to gather features from both images, allowing for constructing more semantically informative and robust features for this query.


\begin{table}[b]
\centering
\begin{minipage}[h]{0.49\textwidth}
\centering
\captionof{table}{\textbf{Results on multi-view map construction.} We apply our view attention to MapTR~\cite{MapTR}.}
\label{tab:hd_map_construction}
\resizebox{\linewidth}{!}{
\begin{tabular}
{l|c|ccc}
\toprule
Method & mAP & AP$_{ped}$ & AP$_{div}$ & AP$_{bound}$ \\
\midrule
MapTR  & 44.73 & 39.69 & 44.59 & 49.91 \\
\rowcolor[RGB]{230,230,230} w/ ViewAttn & \textbf{50.55} & \textbf{45.85} & \textbf{52.82} & \textbf{52.98} \\
\bottomrule
\end{tabular}
}
\end{minipage}
\hfill
\begin{minipage}[h]{0.49\textwidth}
\centering
\caption{\textbf{Results on multi-view 3D object detection.} We apply our view attention to DETR3D~\cite{DBLP:conf/corl/WangGZWZ021}.}
\resizebox{\linewidth}{!}{
  \begin{tabular}
{l|c|c|c|c}
\toprule
Method & mAP$\uparrow$ & NDS$\uparrow$  & mATE$\downarrow$ & mAVE$\downarrow$\\
\midrule
DETR3D  & 0.347 & 0.422 & 0.765 & 0.876 \\
\rowcolor[RGB]{230,230,230} w/ ViewAttn & \textbf{0.388} & \textbf{0.441} & \textbf{0.712} & \textbf{0.874}\\
\bottomrule
\end{tabular}
}
\label{tab:3d_detection}
\end{minipage}
\end{table}

\noindent{\textbf{Application of View Attention.}} We also apply our view attention to other tasks to assess its scalability, including MapTR~\cite{MapTR} for HD map construction and DETR3D~\cite{DBLP:conf/corl/WangGZWZ021} for 3D object detection, both of them collect multi-view image features in a similar way to the projection-first method analyzed in \cref{sec:intro}. We replace their corresponding feature collection module with our view attention. The evaluation metrics are available in their respective literatures. We train MapTR for 24 epochs with an image size of $324 \times 576$, and Res50 as the backbone. \cref{tab:hd_map_construction} indicates our ViewAttn-MapTR achieves a substantial 5.82\% improvement on mAP, demonstrating a remarkable increase in convergence speed. For training DETR3D, we use Res101-DCN as the backbone without CBGS~\cite{Zhu2019ClassbalancedGA}. \cref{tab:3d_detection} shows our ViewAttn-DETR3D also results in 4.1\% and 1.9\% improvements on mAP and NDS respectively. Unlike a considerable amount of recent researches mainly focusing on temporal modeling, and lacking the study of the fundamental spatial interaction for multi-camera 3D perception, our work reveals the limitations of the widely used projection-first method and highlights substantial research opportunities that remain unexplored.

\begin{table}[b]
\caption{\textbf{Ablation study for the streaming temporal attention.} ``w/o TempAttn.'' denotes disabling the streaming temporal attention module.}
\setlength{\tabcolsep}{3.5pt}
\centering
\begin{tabular}
{l|>{\colortable}c|c|c|c|ccc}
\toprule
Method & mIoU & \text{IoU}$_{geo}$ & mAVE$\downarrow$ & FPS$\uparrow$ & car & truck & vege. \\
\midrule
w/o TempAttn. & 39.28 & 68.38 & 1.125 & \textbf{4.2} & 49.97 & 37.38 & 37.27 \\
Voxel to BEV & 41.62 & 71.53 & 1.104 & 3.9 & 52.16 & 40.59 & 40.68 \\
BEV to BEV & \textbf{42.54} & \textbf{72.36} & \textbf{0.412} & 4.1 & \textbf{52.71} & 41.04 & \textbf{41.56} \\
\midrule
Queue $N=1$ & 41.67 & 71.18 & 0.426 & 4.1 & 52.56 & 40.36 & 40.28\\
Queue $N=2$ & 42.28 & 71.95 & 0.421 & 4.1 & 52.53 & 40.75 & 41.11 \\
Queue $N=3$ & 42.42 & 72.18 & 0.415 & 4.1 & 52.70 & \textbf{41.20} & 41.48 \\
\bottomrule
\end{tabular}
\label{tab:streamattn}
\end{table}

\subsubsection{Streaming Temporal Attention.} We train both the occupancy head and the flow head simultaneously to evaluate our temproal modeling in \cref{tab:streamattn}. Overall, our streaming temporal attention improves the mIoU by 3.26 and produces more reasonable flow prediction. Notably, the ``BEV-to-BEV'' interaction method of our streaming temporal attention as depicted in \cref{sec:streaming} outperforms the ``Voxel-to-BEV'' interaction method by 0.92 mIoU, with less computational effort by reducing the number of interaction queries. Due to the presence of a substantial number of empty voxels, compressing voxel queries into BEV queries leads to more pure information and thus enhances the temporal interaction. Additionally, high mAVE indicates that the ``Voxel-to-BEV'' interaction method faces convergence issues in the flow task.

\noindent{\textbf{Length of the Memory Queue.}} We also study the influence of the queue length in \cref{tab:streamattn}, which shows that the performance improves as the queue length increases, proving the effectiveness of our multi-frame interaction mechanism compared to the single-frame interaction mechanism with $N=1$ as BEVFormer~\cite{DBLP:conf/eccv/LiWLXSLQD22}. The performance starts to plateau after $N=3$ as in \cite{Wang_2023_ICCV}. We use $N=4$ in our framework, greater than which the improvement becomes ignorable. As presented in \cref{tab:streamattn} under the ``FPS'' column, thanks to the efficient streaming memory mechanism, the additional time cost brought by multi-frame temporal interaction is negligible.

\subsection{Qualitative Results}

\begin{figure*}[t]
\begin{center}
\includegraphics[width=\linewidth]{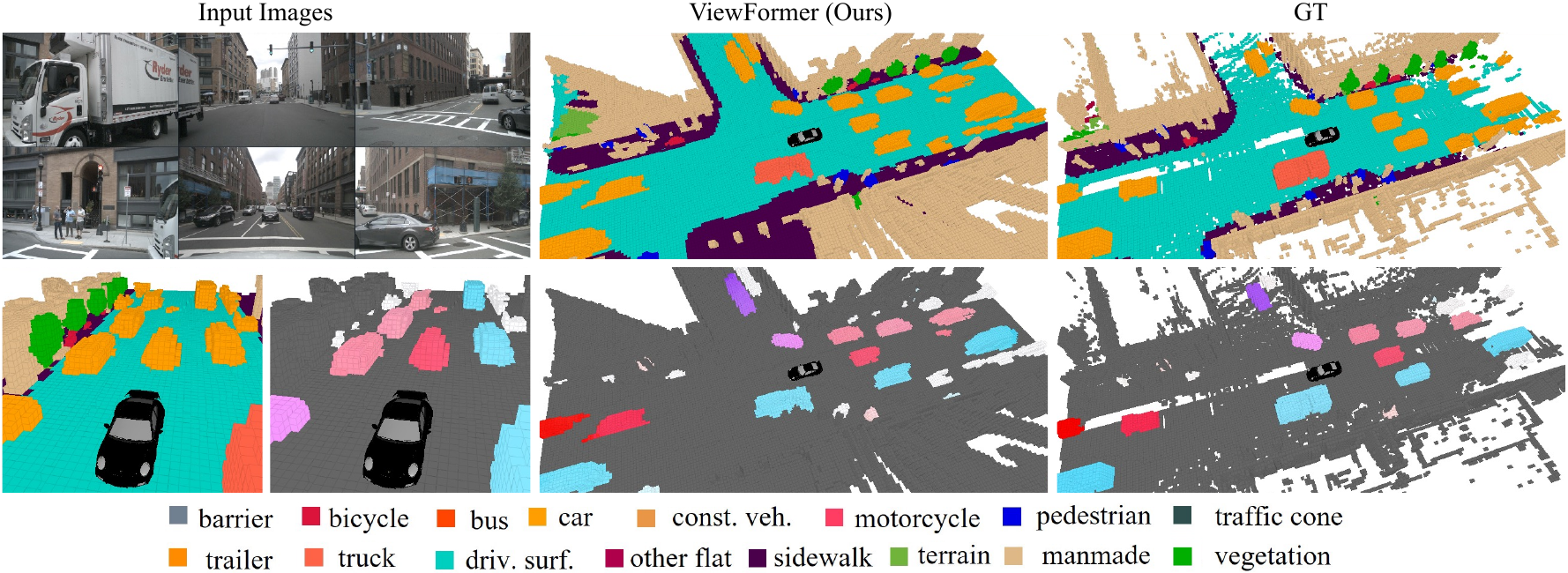}
\end{center}
   \caption{\textbf{Qualitative results of 3D occupancy and occupancy flow prediction.} 3D occupancies are color-coded according to semantic categories. For flow visualization, we utilize color and brightness to represent the flow direction and magnitude respectively, following the convention of the optical flow fields~\cite{DBLP:conf/eccv/TeedD20, DBLP:conf/iccv/BakerSLRBS07}.}
\label{fig:quali_show}
\end{figure*}

\begin{wrapfigure}{r}{0.5\textwidth}
   \centering
    \includegraphics[width=0.5\textwidth]{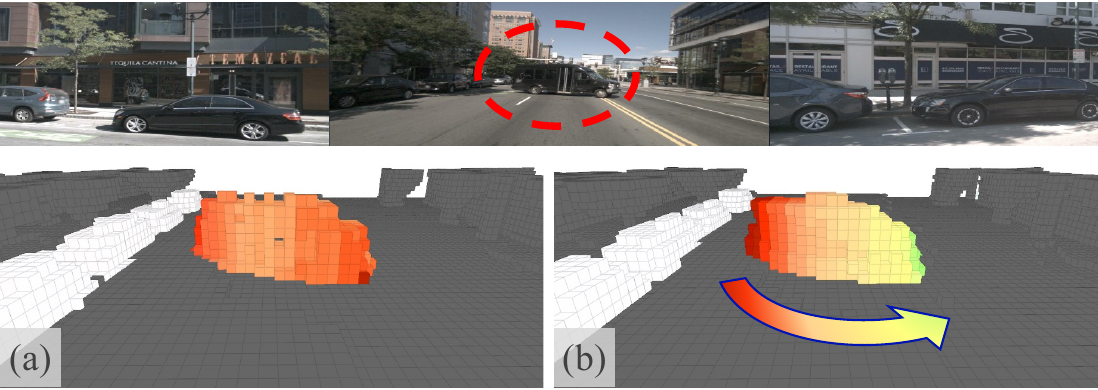}
  \caption{\small \textbf{Visualization on flow predictions.} (a) Flow predictions supervised by object-level flow. (b) Flow predictions supervised by our FlowOcc3D.
  }
  \label{fig:flow_ob_occ}
\end{wrapfigure}

We present qualitative results of our ViewFormer in \cref{fig:quali_show}. An additional animated video is also provided in the supplementary material. Moreover, we demonstrate a qualitative comparison for flow predictions supervised by object-level flow (a) and occupancy-level flow (b) in \cref{fig:flow_ob_occ}, the model trained with our occupancy-level FlowOcc3D dataset achieves more reasonable results for a turning car, providing fine-grained motion information for AD systems.

%% file: sec/6_conclusion.tex
\section{Conclusion}

We present the ViewFormer framework for 3D occupancy and occupancy flow prediction, featuring our proposed view attention that addresses the limitations of the existing projection-first spatial interaction method, as well as the streaming temporal attention designed for multi-frame temporal interaction. Furthermore, we build a novel occupancy-level flow benchmark FlowOcc3D to explore the potential of occupancy flow representation for dynamic scenes, which we also contribute to the research community. Our approach demonstrates significant advancements over previous state-of-the-art methods.


%% file: sec/X_suppl.tex
\clearpage
\setcounter{page}{1}

\section{Appendix}
\label{sec:appendix}

\subsection{Convergence Analysis for View Attention}


\begin{figure}[h]
\begin{center}
\includegraphics[width=\linewidth]{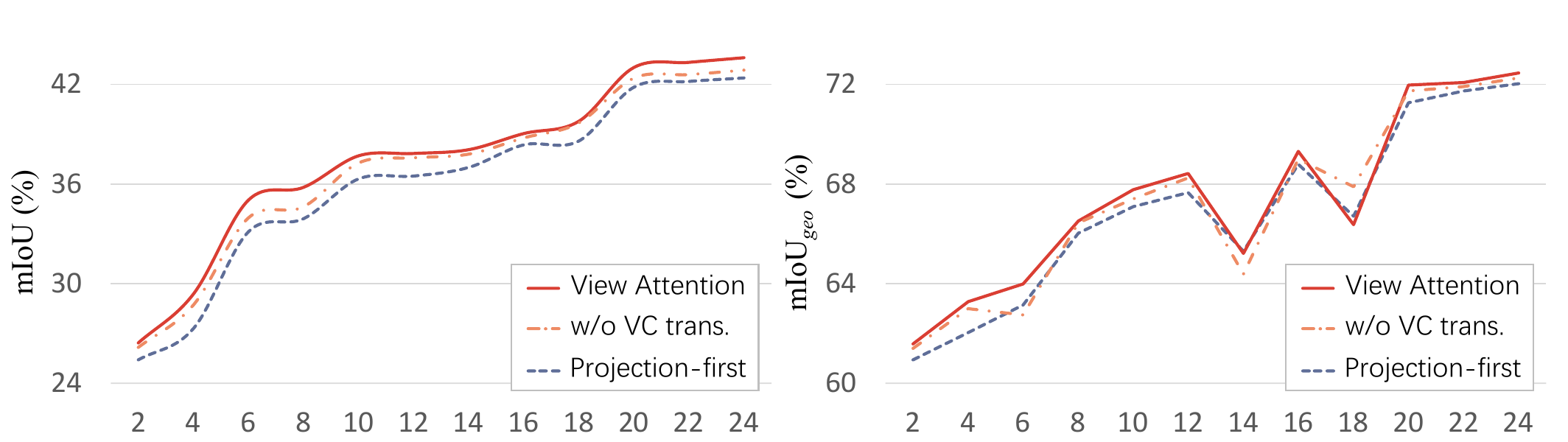}
\end{center}
   \caption{\textbf{Convergence Analysis.} Our view attention achieves not only faster convergence but also impressive performance. All models are trained for 24 epochs.}
\label{fig:convergence_exp}
\end{figure}

\noindent \cref{fig:convergence_exp} illustrates the convergence analysis of the spatial interaction module, where we replace our view attention with the projection-first interaction method as BEVFormer~\cite{DBLP:conf/eccv/LiWLXSLQD22}. The results showcase that our view attention achieves not only faster convergence but also impressive performance. The term "w/o VC trans." denotes learning 3D sample points in the ego-centric perception coordinate system without view angle rotation, highlighting the insightful rotation invariance introduced by our view attention.

\subsection{Comparison of Temporal Modeling}

\begin{table}[h]
\caption{\textbf{Comparison of Temporal Modeling.} ``\textbf{SW}'' denotes sliding window training approach.}
\setlength{\tabcolsep}{3.5pt}
\centering
\begin{tabular}{l|c|c|c|c|c}
\toprule
 Operation & Train & Time (h) & mIoU & \text{IoU}$_{geo}$ & mAVE$\downarrow$ \\
\midrule
DeformAttn. & \textbf{SW} & 36 & 42.17 & 72.14 & 0.417 \\
DeformAttn. & Video & 13 & \textbf{42.54} & \textbf{72.36} & 0.\textbf{412} \\
CNN & Video & 13 & 41.95 & 71.73 & 0.424 \\
\bottomrule
\end{tabular}
\label{tab:videoSW}
\end{table}

\noindent ViewFormer is trained and tested with online streaming video, while the baseline \cite{DBLP:conf/eccv/LiWLXSLQD22} is trained with a local sliding window and tested with online streaming video. The results in \cref{tab:videoSW} demonstrate that maintaining consistency between training and inference leads to improved accuracy. It is noteworthy that the sliding window training approach is time-consuming due to redundant re-inference for each frame window. Benefiting from the streaming memory mechanism~\cite{Wang_2023_ICCV, Park2022TimeWT}, our ViewFormer pushes each frame into the memory queue for subsequent temporal interaction, thus significantly reducing the training time. We also replace our transformer-based operation, {\em i.e.} the deformable attention~\cite{DBLP:conf/iclr/ZhuSLLWD21}, with a CNN-based temporal interaction operation as in \cite{Li2023FBBEV,Park2022TimeWT}, which yields suboptimal results.

\subsection{Depth Supervision Analysis}

\begin{table*}[h]
\caption{\textbf{Depth Supervision Analysis for 3D Occupancy Prediction on Occ3D benchmark.}
}
\small
\setlength{\tabcolsep}{2.0pt}
\centering
\resizebox{\textwidth}{!}{

\begin{tabular}{@{}l|c|c|>{\colortable}c|c|ccccccccccccccccc@{}}
\toprule
Method                  & Backbone & Depth & mIoU & \text{IoU}$_{geo}$  &\rotatebox{90}{others} &\rotatebox{90}{barrier} & \rotatebox{90}{bicycle} &  \rotatebox{90}{bus}   & \rotatebox{90}{car}   & \rotatebox{90}{const. veh.} & \rotatebox{90}{motorcycle} & \rotatebox{90}{pedestrian} & \rotatebox{90}{traffic cone} & \rotatebox{90}{trailer} & \rotatebox{90}{truck} & \rotatebox{90}{driv. surf.} & \rotatebox{90}{other flat} & \rotatebox{90}{sidewalk} & \rotatebox{90}{terrain} & \rotatebox{90}{manmade} & \rotatebox{90}{vegetation} \\
\midrule
FB-OCC~\cite{Li2023FBBEV} & InternT & \Checkmark & 38.69 & 69.95 & 11.4 & 41.42 & 24.27 & 46.01 & 49.38 & \textbf{24.56} & 27.06 & 28.09 & 25.61 & 32.23 & 38.46 & 80.97 & 42.99 & 50.95 & 56.15 & 40.55 & 37.61 \\
\rowcolor[RGB]{230,230,230}ViewFormer (ours) & InternT & \Checkmark & \textbf{43.61} & \textbf{72.46} & \textbf{13.82} & \textbf{50.32} & \textbf{29.49} & \textbf{49.24} & \textbf{54.52} & 24.34 & \textbf{32.72} & \textbf{31.09} & \textbf{31.49} & \textbf{34.44} & \textbf{41.62} & \textbf{85.47} & \textbf{51.27} & \textbf{59.03} & \textbf{62.15} & \textbf{48.33} & \textbf{42.06} \\
\midrule

FB-OCC~\cite{Li2023FBBEV} & InternT & \XSolidBrush & 36.26 & 67.48 & 10.71 & 37.06 & 23.55 & 42.60 & 47.16 & 19.10 & 25.94 & 25.75 & 23.52 & 29.91 & 35.60 & 79.73 & 41.41 & 49.51 & 54.50 & 35.58 & 34.23 \\
\rowcolor[RGB]{230,230,230}ViewFormer (ours) & InternT & \XSolidBrush & \textbf{41.76} & \textbf{71.37} & \textbf{12.98} & \textbf{47.70} & \textbf{27.85} & \textbf{45.12} & \textbf{53.02} & \textbf{21.67} & \textbf{28.93} & \textbf{28.80} & \textbf{29.67} & \textbf{31.02} & \textbf{39.21} & \textbf{86.06} & \textbf{51.22} & \textbf{58.94} & \textbf{62.00} & \textbf{45.50} & \textbf{40.18} \\

\bottomrule
\end{tabular}
}
\label{tab:occ_compe_nodepth}
\end{table*}

\noindent As mentioned in \cref{sec:mainRS} of our paper, we follow FB-OCC~\cite{Li2023FBBEV} to adopt depth supervision in backbone training for a fair comparison (note that depth supervision is not adopted in \cref{tab:occ_method}). Now, let's delve into the detailed effects of depth supervision. In \cref{tab:occ_compe_nodepth}, we present 3D occupancy results with depth supervision disabled. In the absence of depth supervision, the performance gap between our ViewFormer and FB-OCC widens: our ViewFormer outperforms FB-OCC by 5.5 mIoU and 3.89 mIoU$_{geo}$. It is evident that without depth supervision, our model shows a modest drop of 1.09 mIoU$_{geo}$ (72.46 vs. 71.37), while FB-OCC presents a larger drop of 2.47 mIoU$_{geo}$ (69.95 vs. 67.48). Due to the limited availability of depth data in vision-centric datasets, our method, with lower reliance on depth information, demonstrates superior generality.

\subsection{Two-DOF View Attention}

\begin{figure}[h]
\begin{center}
\includegraphics[width=0.6\linewidth]{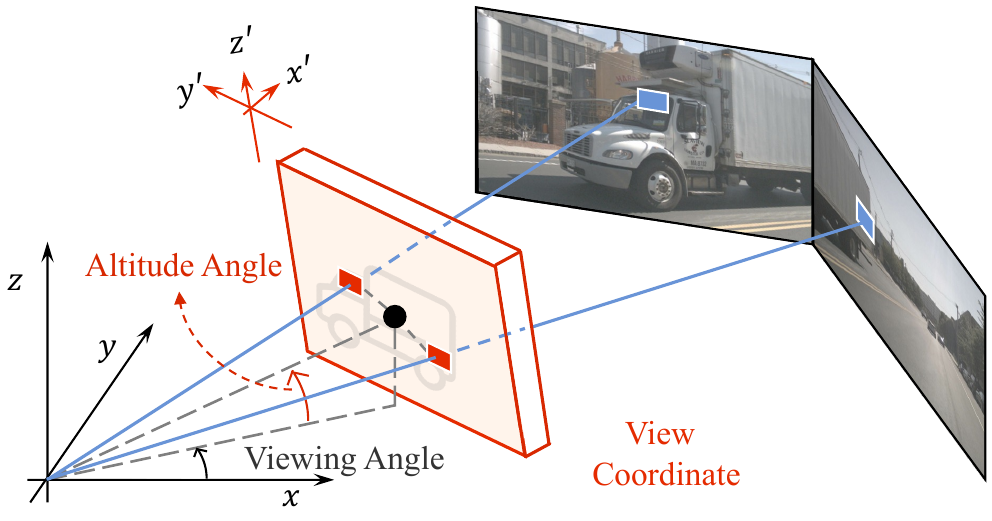}
\end{center}
   \caption{Two-DOF view attention dealing with both the view angle and altitude angle.}
\label{fig:tp_altitude}
\end{figure}

\noindent In \cref{sec:viewattn}, we present a single degree of freedom (DOF) version view attention, correlated to a query's view angle. As illustrated in \cref{fig:tp_altitude}, our two-DOF view attention involves an additional altitude angle. Similar to \cref{eq:eq2}, the query-related altitude angle $\boldsymbol{\phi}$ and the corresponding rotation matrix $\boldsymbol{R}(\phi)$ around the y-axis are calculated as:
\begin{equation}
\label{eq:eq8}
\begin{aligned}
\phi &=  \text{arctan2}\left(z, \sqrt{x^2 + y^2} \right),\\
\boldsymbol{R}(\phi) &= 
 \left[ \begin{array}{ccc}
    \cos\phi & 0 & -\sin\phi\\
    0 & 1 & 0\\
    \sin\phi & 0 & \cos\phi\\
  \end{array}
  \right].\\
\end{aligned}
\end{equation}

Dealing with both the view angle and altitude angle, \cref{eq:eq3} can be extended as:

\begin{equation}
    \boldsymbol{p_s}
    = \boldsymbol{p}+\boldsymbol{R}(\theta)\cdot\boldsymbol{R}(\phi)\cdot \Delta\boldsymbol{p}^\mathcal{T}.
\label{eq:eq9}
\end{equation}

In contrast to our single-DOF version dealing with only the view angle, the two-DOF version actually brings no performance gain. The reason is considered to lie in the fact that like the nuScenes~\cite{DBLP:conf/cvpr/CaesarBLVLXKPBB20,DBLP:journals/ral/FongMHZCBV22} dataset, the cameras are almost horizontally mounted in an AD system, our single-DOF version is therefore capable of representing the rotation invariance. In other scenarios such as the field of 3D reconstruction where the cameras are irregularly distributed, however, the two-DOF version presents a valuable option.

\subsection{Discussion about the Necessity of Occupancy Flow}

In our paper, we create our occupancy flow benchmark FlowOcc3D, to explore the potential of fine-grained representation of dynamic scenes. Questions may arise regarding how the integrity of objects is maintained in such representation. We would like to emphasize that the occupancy-level flow representation tackles challenges faced in the traditional object-level flow representation, {\em e.g.}, lack of effective representation for irregularly shaped objects and background information. Existing occupancy tasks do not address the object integrity issue, but rather resembles semantic segmentation in 3D space. The object integrity is more similar to the task handled in instance segmentation, which is in fact out the scope of occupancy tasks.